\pdfoutput=1

\documentclass[11pt]{article}

\usepackage[final]{emnlp2021}

\usepackage{latexsym}

\usepackage{savesym}
\usepackage{amsmath}
\savesymbol{iint}
\usepackage{txfonts}
\restoresymbol{TXF}{iint}

\DeclareMathOperator*{\argmax}{arg\,max}

\usepackage{graphicx}
\usepackage{multirow}

\usepackage[T1]{fontenc}

\usepackage[utf8]{inputenc}

\usepackage{microtype}

\title{Position-Invariant Truecasing with a Word-and-Character Hierarchical Recurrent Neural Network}

\author{Hao Zhang \and You-Chi Cheng \and Shankar Kumar \and Mingqing Chen \and  Rajiv Mathews \\
Google \\
  \texttt{\{haozhang,youchicheng,shankarkumar,mingqing,mathews\}@google.com}}

\begin{document}
\maketitle
\begin{abstract}
Truecasing is the task of restoring the correct case (uppercase or lowercase) of noisy text generated either by an automatic system for speech recognition or machine translation or by humans. It improves the performance of downstream NLP tasks such as named entity recognition and language modeling. We propose a fast, accurate and compact two-level hierarchical word-and-character-based recurrent neural network model, the first of its kind for this problem. Using sequence distillation, we also address the problem of truecasing while ignoring token positions in the sentence, i.e. in a position-invariant manner.
\end{abstract}

\section{Introduction}
\label{sec:intro}

Automatically generated texts such as speech recognition (ASR) transcripts as well as user-generated texts from mobile applications such as Twitter Tweets \cite{DBLP:conf/www/NebhiBG15} often violate the grammatical rules of casing in English and other western languages \cite{grishina-etal-2020-truecasing}. The process of restoring the proper case, often known as \textbf{t{R}u{E}cas{I}ng} \cite{lita-etal-2003-truecasing}, is not only important 
for the ease of consumption by end-users (e.g. in the scenario of automatic captioning of videos), but is also crucial to the success of downstream NLP components. For instance, named entity recognition (NER) relies heavily on the case features \cite{Mayhew_Nitish_Roth_2020}. In language modeling (LM), truecasing is crucial for both pre- and post-processing. It has been used for restoring case and improving the recognition performance for ASR ({\em post-processing}, \citealp{4960690, 6638968}) as well as case normalization of user text prior to LM training ({\em pre-processing}, \citealp{DBLP:conf/conll/ChenSMWABR19}). The latter employs Federated Learning \cite{DBLP:conf/aistats/McMahanMRHA17}, a privacy-preserving learning paradigm, on distributed devices. The need for normalizing text on a variety of mobile devices makes it an imperative to develop a fast, accurate and compact truecaser.

From a modeling perspective, there has been a divide between character-based and word-based approaches.
Most of the earlier works are word-based \cite{lita-etal-2003-truecasing, chelba-acero-2004-adaptation}.
In the word-based view, the task is to classify each word in the input into one of the few classes \cite{lita-etal-2003-truecasing}: all lowercase (\texttt{LC}), first letter uppercase (\texttt{UC}), all letters uppercase (\texttt{CA}), and mixed case (\texttt{MC}). The main drawback of the word-based approach is that it does not generalize well on unseen words and mixed case words.
The development of character-based neural network modeling techniques \cite{pmlr-v32-santos14} led to the introduction of character-based truecasers \cite{susanto-etal-2016-learning, 9031494}.
In the character-based view, the task is to classify each character in the input into one of the two classes: \texttt{U} and \texttt{L}, for uppercase and lowercase respectively.
The shortcoming of character-based models is inefficiency. As word-level features are always important for classification, character-based models need to be deep enough to capture high-level features, which exacerbates their slowness. 

We view truecasing as a special case of text normalization \cite{DBLP:journals/coling/ZhangSNSPGR19} with similar trade-offs for accuracy and efficiency. Thus, we classify input words into one of two classes: \texttt{SELF} and \texttt{OTHER}. Words labeled \texttt{SELF} will be copied as-is to the output. Words labeled \texttt{OTHER} will be fed to a sub-level character-based neural network along with the surrounding context, which is encoded. The task of the sub-network to perform non-trivial transduction such as:
\texttt{iphone} $\rightarrow$ \texttt{iPhone},
\texttt{mcdonald's} $\rightarrow$ \texttt{McDonald's} 
and \texttt{hewlett-packard} $\rightarrow$  \texttt{Hewlett-Packard}.
The two-level hierarchical network is fast while also being accurate.

\newcite{lita-etal-2003-truecasing} highlighted the uneventful uppercasing of the first letter in a sentence, and informally defined a new problem: ``{\em a more interesting problem from a truecasing perspective is to learn
how to predict the correct case of the first word in a
sentence (i.e. not always UC)}''. We address the problem with a modified sequence distillation \cite{kim-rush-2016-sequence} method. The intuition is inspired by the distinction between {\em positional} and {\em intrinsic} capitalization made by \newcite{6638968}. We first train a teacher model on positionally capitalized data and then distill sub-sequences from the teacher model when training the student model. By doing so, the student model is only exposed to intrinsically capitalized examples. The method accomplishes both model compression and the position-invariant property which is desirable for mobile NLP applications.

Section~\ref{sec:related} discusses related work. Section~\ref{sec:problem_and_model} defines the machine learning problem and presents the word- and character-based hierarchical recurrent neural network architecture. Section~\ref{sec:position_invariant} introduces the modified sequence distillation method for positional invariant truecasing, as well as a curated Wikipedia data set for evaluation\footnote{https://github.com/google-research-datasets/wikipedia-intrinsic-capitalization}. Section~\ref{sec:expts} shows the results and analysis.

\section{Related Work}
\label{sec:related}
Word-based truecasing has been the dominant approach for a long time since the introduction of the task by \newcite{lita-etal-2003-truecasing}.
Word-based models can be further categorized into generative models such as HMMs \cite{lita-etal-2003-truecasing, 4960690, 6638968, DBLP:conf/www/NebhiBG15} and
discriminative models such as Maximum-Entropy Markov Models \cite{chelba-acero-2004-adaptation}, Conditional Random Fields \cite{DBLP:conf/naacl/WangKM06}, and most recently Transformer neural network models \cite{9041202, 10.1007/978-3-030-50146-4_52, sunkara-etal-2020-robust}.
Word-based models need to refine the class of mixed case words because there is a  combinatorial number of possibilities of case mixing for a word (e.g., \texttt{LaTeX}). \newcite{lita-etal-2003-truecasing, chelba-acero-2004-adaptation} suggested using either all of the forms or the most frequent form of mixed case words in the training data. The large-scale finite state transducer (FST) models by \newcite{6638968} used all known forms of mixed case words to build the ``capitalization'' FST. \newcite{DBLP:conf/naacl/WangKM06} used a heuristic \textbf{GEN} function to enumerate a superset of all forms seen in the training data.
But others \cite{9041202, 10.1007/978-3-030-50146-4_52} have chosen to simplify the problem by mapping mixed case words to first letter uppercase words. \newcite{sunkara-etal-2020-robust} only evaluated word class F1, without refining the class of \texttt{MC}.

Character-based models have been explored largely after the dawn of modern neural network models. \newcite{susanto-etal-2016-learning} first introduced character-based LSTM for this task and completely solved the mixed case word problem. Recently, \newcite{grishina-etal-2020-truecasing} compared character-based $n$-gram ($n$ up to 15) language models with the character LSTM of \newcite{susanto-etal-2016-learning}.
\newcite{9031494} advanced the state of the art with a character-based CNN-LSTM-CRF model which introduced local output label dependencies.

Text normalization is the process of transforming text into a canonical form. Examples of text normalization include but are not limited to written-to-spoken text normalization for speech synthesis \cite{DBLP:journals/coling/ZhangSNSPGR19}, spoken-to-written text normalization for speech recognition \cite{Peyser2019}, social media text normalization \cite{10.1145/2414425.2414430}, and historical text normalization \cite{makarov-clematide-2020-semi}. Truecasing is a problem that appears in both spoken-to-written and social media text normalization.

\section{Formulation and Model Architecture}
\label{sec:problem_and_model}
The input is a sequence of all lowercase words $\vec{\mathrm{X}}=(x_1,\dots,x_l)$. The output is a sequence of words with proper casing $\vec{\mathrm{Y}}=(y_1,\dots,y_l)$. We introduce a latent sequence of class labels $\vec{\mathrm{C}}=(c_1,\dots,c_l)$, where $c_i \in \{\texttt{S=SELF,O=OTHER}\}$. We also use the notation $x_i^j$ and $y_i^j$ to represent the $j\text{-th}$ character within the $i\text{-th}$ word.

The model is trained to predict the probability: 
\begin{equation}
\label{eqn:hier}
    \mathrm{P}(\vec{\mathrm{Y}}|\vec{\mathrm{X}}) = \sum_{\vec{\mathrm{C}}}\mathrm{P}(\vec{\mathrm{Y}}|\vec{\mathrm{X}}, \vec{\mathrm{C}}) \cdot \mathrm{P}(\vec{\mathrm{C}}|\vec{\mathrm{X}}),
\end{equation}
where $\mathrm{P}(\vec{\mathrm{C}}|\vec{\mathrm{X}})$ is a word-level model that predicts if a word should stay all lowercase (\texttt{SELF}) or change to a different case form (\texttt{OTHER}), which assumes label dependency between $c_1,\dots,c_{i-1}$ and $c_i$.
\begin{equation}
\label{eqn:word_based}
    \mathrm{P}(\vec{\mathrm{C}}|\vec{\mathrm{X}})  =  \prod_{i=1}^{l} {\mathrm{P}(c_i|c_1,\dots,c_{i-1},\vec{\mathrm{X}}})
\end{equation}
The label sequence $\vec{\mathrm{C}}$ works as a gating mechanism,
\begin{align}
\label{eqn:dispatch}
    \mathrm{P}(\vec{\mathrm{Y}}|\vec{\mathrm{X}}, \vec{\mathrm{C}}) =& \prod_{i=1}^{l}{\delta(c_i, \texttt{O})  \mathrm{P}(y_i|{\mathrm{X}) + \delta(c_i, \texttt{S})} \delta(x_i, y_i)},
\end{align}
where $\mathrm{P}(y_i|{\mathrm{X}})$ is a character-level model that predicts each output character, assuming dependency between characters within a word: $y_i^{1},\dots,y_i^{j-1}$ and $y_i^j$, but no cross-word dependency between $y_1,\dots,y_{i-1}$ and $y_i$, and \texttt{S} and \texttt{O} denote \texttt{SELF} and \texttt{OTHER} respectively.
\begin{equation}
\label{eqn:char_based}
    \mathrm{P}(y_i|{\mathrm{X}}) = \prod_{j=1}^{j=|x_i|}{\mathrm{P}(y_i^j|y_i^{1},\dots,y_i^{j-1},\vec{\mathrm{X}})}
\end{equation}
Given that $\delta(c_i, \texttt{S}) \equiv \delta(x_i, y_i)$, 
we can derive the log likelihood of Equation~\ref{eqn:hier} as:
\begin{align}
    \log(\mathrm{P}(\vec{\mathrm{Y}}|\vec{\mathrm{X}})) &= \sum_{i=1}^{l}{\delta(c_i, \texttt{O}) \log(\mathrm{P}(y_i|{\mathrm{X}}))} \nonumber \\
    &+\log(\mathrm{P}(\vec{\mathrm{C}}|\vec{\mathrm{X}}))
\end{align}
Equation~\ref{eqn:word_based} and Equation~\ref{eqn:char_based} are sequence-to-sequence (seq2seq) problems. Unlike the general machine translation problem with unequal number of input and output tokens which requires a soft attention mechanism \cite{DBLP:journals/corr/BahdanauCB14}, both of our seq2seq problems can assume hard alignment between the output label at each time step and the input symbol at the same time step ($c_i$ is aligned to $x_i$, $y_i^j$ is aligned to $x_i^j$).

\begin{figure*}[hbt!]
  \includegraphics[width=\textwidth]{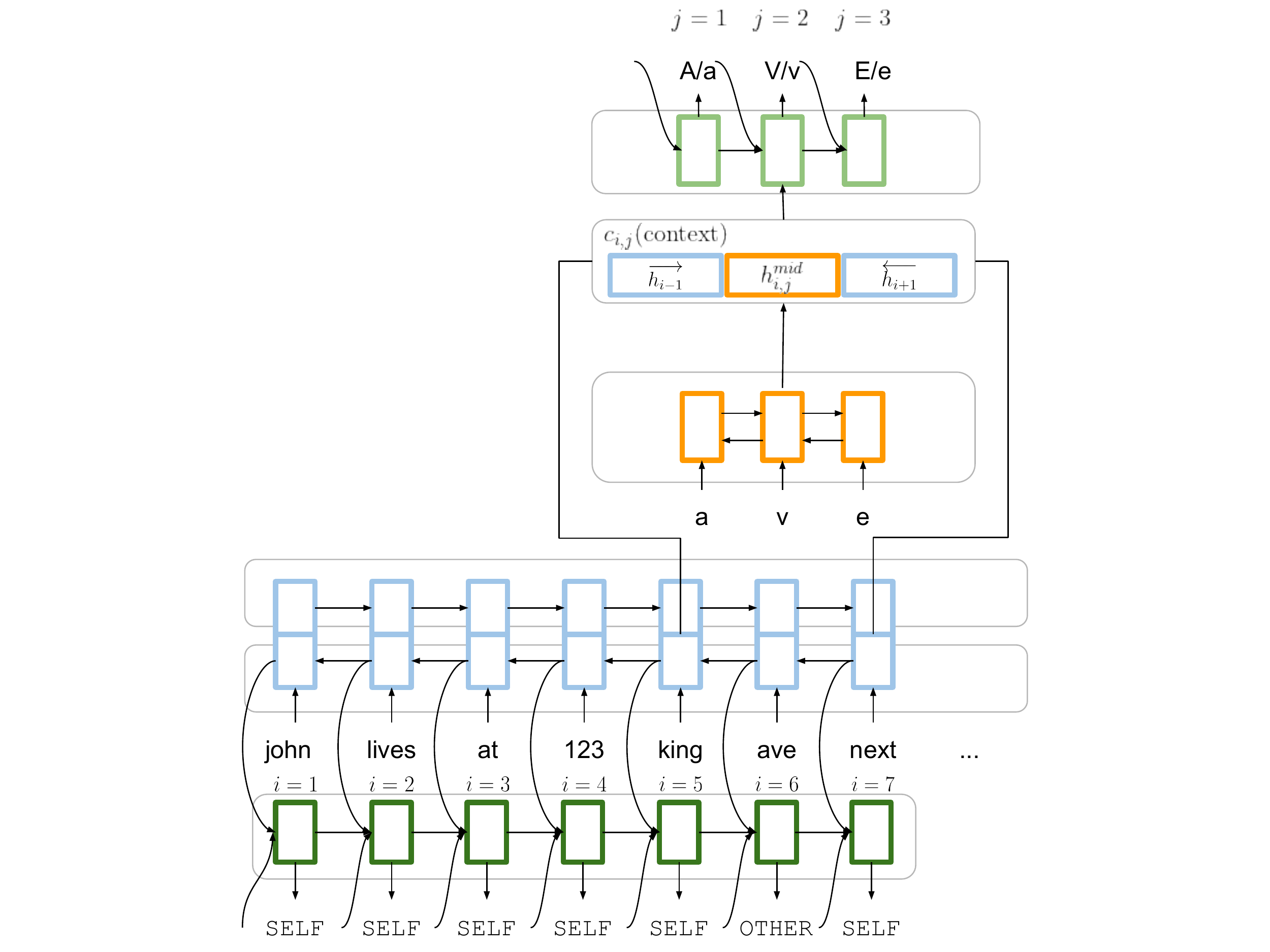}
\caption{\label{fig:architecture} Hierarchical RNN architecture.} 
\end{figure*}

We follow the multi-task recurrent neural network architecture of \newcite{DBLP:journals/coling/ZhangSNSPGR19}.
Figure~\ref{fig:architecture} displays the rolled-out network of the two-level hierarchical RNN on a typical input sentence. The key difference with this work is that in our setting, the models at both levels perform sequence tagging whereas their second-level model is a full-fledged sequence-to-sequence model with soft attention \cite{DBLP:journals/corr/BahdanauCB14}.

The exact inference requires searching over all possible $\vec{\mathrm{C}}$ and  $\vec{\mathrm{Y}}$ which is infeasible for RNNs. But we can do beam search for both Equation~\ref{eqn:word_based} and Equation~\ref{eqn:char_based} and and use either the entire beam of hypotheses of $\vec{\mathrm{C}}$ or simply the best one in Equation~\ref{eqn:dispatch}. In our experiments, using the entire beam only helped slightly.

\section{Position Invariant Modeling}
\label{sec:position_invariant}

In English, the first character in a sentence is usually capitalized. This type of capitalization is done for a different reason from the one used for capitalization of words which are parts of named entities. 
\newcite{6638968} called the first type {\em positional} and the second type {\em intrinsic}. A word appearing at the beginning position of a sentence can be capitalized due to either one of the two reasons. But it is more interesting to recover the intrinsic capitalization which is independent of word positions. For example, in \texttt{apples are nutritious}, the intrinsic capitalization form of the word \texttt{apples} is lowercase. But in \texttt{Apple is a personal computer company}, the first letter of \texttt{Apple} should be capitalized.

A truecasing model that recovers the position-invariant word forms can benefit downstream NLP components such as NER and LM models. Without position-invariant truecasing, almost every word that can appear at the beginning of a sentence will have two forms in the corpus.
Normalizing out the positional factor can reduce data sparsity and reduce the vocabulary size for word-based LMs and sub-word-based LMs (to a smaller extent).

\newcite{6638968} provided a simple solution. Their FST-based truecaser recovers positional capitalization. However, at inference time they add a vacuous word such as \texttt{and} or the \texttt{<UNK>} symbol standing for unknown words in the vocabulary to the beginning of any input sentence. The case recovered for the vacuous prefix word is unimportant. But because of the prefix, it is more likely the intrinsic capitalization of the original sentence including the first word will be recovered.

We combine this technique with the sequence knowledge distillation technique of \newcite{kim-rush-2016-sequence}. Mathematically speaking, we use $(\vec{\mathrm{X}}, \hat{\mathrm{Y}})$ as training examples where $\hat{\mathrm{Y}}$ is obtained from inference using the teacher model, which is trained for positional truecasing using $(\vec{\mathrm{X}}, \vec{\mathrm{Y}})$.

\begin{align}
\label{eqn:distill}
\hat{\mathrm{Y}} = \argmax_{\mathrm{Y}'}{\mathrm{P}(\mathrm{Y_{prefix}} \oplus \mathrm{Y}'|\mathrm{X_{prefix}} \oplus \vec{\mathrm{X}})}
\end{align}

While it is possible to explore a set of prefixes in Equation~\ref{eqn:distill}, in practice, using a single common prefix \texttt{so,} works well. In Section~\ref{sec:distill_prefixes}, we examine the effect of prefix choices on the accuracy of the model.

The knowledge distillation technique is effective for up-training a smaller model with fewer parameters. Thus, the partial sequence distillation method outlined here achieves two goals: model compression and position invariant truecasing.

\subsection{Intrinsically Capitalized Data}
\label{sec:wiki_edit_data}
We use sequence distillation to train models. However, evaluating truecasing independent of word positions requires human-curated data. We resort to crowd-sourcing. In the annotation task, annotators are asked to provide intrinsic capitalization of the input sentence following instructions and examples. They are also required to correct any other casing errors in the data. We use Wikipedia edit history data \cite{lichtarge-etal-2019-corpora} for the purpose.


The crowd-sourcing data was sampled in the following way.
First, we picked the top 55K sentences with the lowest perplexity scores according to an n-gram language model.
Next, we filtered the data so that there is a 75\% disagreement rate between an in-house large-scale FST truecaser and the neural truecasing model. 
Finally, we removed sentences containing quotation marks because the capitalization rules within the quoted sentences tend to be quite different from the other sentences.
We assigned three annotators to each sentence. If all three annotators were in agreement, we chose the agreed-upon version. Otherwise, we decided which annotation to use. The final data set has 1,200 sentences in total.

\section{Experiments}
\label{sec:expts}
All models take lowercase tokenized sentences as input and output the same tokens with predicted casing. The predicted tokens are compared against references.
Following \newcite{susanto-etal-2016-learning}, we use non-lowercase (NL) F1 as the metric for comparison.
\begin{align*}
\text{NL Precision} = &\frac{\text{\# of correct NL predictions}}{\text{\# of NL predictions}} \\
\text{NL Recall} = &\frac{\text{\# of correct NL predictions}}{\text{\# of NL references}}
\end{align*}
NL F1 is the harmonic mean of NL Precision and Recall. For brevity, we use precision, recall and F1 without qualification.

Our models as depicted in Figure~\ref{fig:architecture} have a pair of forward and backward RNNs to encode word-level context. The lengths of both RNNs after rolling out are equal to the number of tokens (words) in the input sentence. Words are first encoded based on subword features, namely character $n$-grams with $n$ up to 3 \cite{DBLP:journals/coling/ZhangSNSPGR19}. In the example of Figure~\ref{fig:architecture}, the character $n$-grams representing the word \texttt{ave} include \texttt{a}, \texttt{v}, \texttt{e}, \texttt{<s>a}, \texttt{av}, \texttt{ve}, \texttt{e<s>}, \texttt{<s>av}, \texttt{ave}, and \texttt{ve<s>}, where \texttt{<s>} is the placeholder symbol for the word beginning and ending positions. These sparse $n$-gram features are mapped to a dense linear space of $[0, \text{\# of buckets})$ with a hash function \cite{bohnet-2010-top}, where the number of buckets is 5k in our experiments. Each word is then represented as the summation of the embeddings of the hashed character $n$-gram features. The sub-word feature representation is critical to obtain a trade-off between speed and accuracy, as shown in the systematic comparison for the written-to-spoken text normalization task by \newcite{DBLP:journals/coling/ZhangSNSPGR19}.
The output of the context RNNs are used in a multi-task manner for both the word-tagging RNN of Equation~\ref{eqn:word_based} and the character-predicting RNN of Equation~\ref{eqn:char_based} shown as the bottom and the top RNNs in Figure~\ref{fig:architecture}, respectively. Figure~\ref{fig:architecture} has only one word that activates the character-predicting RNN. In reality, there will be the same number of character-level RNN sub-model instances as the number of \texttt{OTHER} classes predicted by the word-level RNN sub-model. We use GRU cells in all these RNNs by default because they allow us to obtained a better trade-off between speed and accuracy when compared to LSTMs.

As Section~\ref{sec:position_invariant} indicated, we train a large teacher model on sentences with positional capitalization which is then distilled to a small student model which learns position-invariant capitalization. The hyper-parameters of both models are listed in Table~\ref{tab:network_specs}.
\begin{table}[ht]
    \centering
    \begin{tabular}{lrr}
                                & teacher & student \\                               
                                & (large) & (small) \\
             \hline
             input embedding size & 512 & 128 \\
             output embedding size & 512 & 128 \\
            \# of forward enc. layers & 2 & 1\\
             \# of backward enc. layers & 2 & 1\\
             \# of dec. layers & 2 & 1\\
             \# of enc. RNN cells & 512 & 128\\
             \# of dec. RNN cells & 512 & 128 \\
             max char $n$-gram order & 3 & 3\\
             buckets of char $n$-grams & 5000 & 5000 \\
             beam size & 2 & 2\\
             \hline
    \end{tabular}
    \caption{Network hyper-parameters.}
    \label{tab:network_specs}
\end{table}

Given that our model has two incarnations of the encoder-decoder architecture for the word- and character-level sub-models, the same set of parameters: input/output embedding size, the number of forward/backward encoder layers, the number of decoding layers, and the number of RNN cells per layer, apply to both sub-models. In Section~\ref{sec:large_scale}, we will describe our implementation of character-only models, which have a single encoder-decoder architecture with the same hyper-parameter sets.

\subsection{Comparison with Published Character-based Neural Models}
\label{sec:char_wiki_simpl}
The character-based models of \newcite{susanto-etal-2016-learning, 9031494} are directly comparable. Both have reported results on the the Wikipedia text simplification data set \cite{coster-kauchak-2011-simple}. 
This is a small data set with 2.9M words for training, 294K words for validation, and 32K words for testing.
The sentences are {\em positionally capitalized} as they originally appear on Wikipedia pages.
We use the same data splits for training/testing. We report word-level precision and recall and F1 for words that are not all lowercase. 
We use the large network specified in Table~\ref{tab:network_specs} and vary the RNN cell type from GRU to LSTM. Following the prior work, we use a learning rate (0.03) and apply an input dropout rate of 0.25 to every RNN cell.
The results on the test set are shown in Table~\ref{tab:wiki_simpl}. When using LSTM cells, the F1 score is higher than the large character LSTM model of \newcite{susanto-etal-2016-learning} by 0.1 and about 0.7 point lower than the CNN-LSTM-CRF model of \newcite{9031494}. 
We speculate that the the hashed character $n$-gram features require careful hyper-parameter search including hashing function selection. It is interesting to experiment with CNNs for sub-word feature representation \cite{pmlr-v32-santos14}.
Despite the differences, the fact that the character-based models do not need any feature engineering like the sub-word features we have is indeed an advantage. In the next section, we will scale up various models and look into the efficiency aspect as well.

\begin{table*}[htb]
    \centering
    \begin{tabular}{cc|c|c|c}
      \multicolumn{2}{c}{\textbf{System}} & \textbf{Precision} & \textbf{Recall} & \textbf{F1} \\
      \hline
      \newcite{susanto-etal-2016-learning}  & LSTM& 93.72 & 92.67 & 93.19 \\
      \newcite{9031494} & CNN-LSTM-CRF& 94.92 & 93.14 & 94.02 \\
      \hline
      \multirow{2}{*}{\textbf{this work}} &GRU & 93.69 & 92.34 & 93.01 \\
     & LSTM & 94.25 & 92.40 & 93.33 \\
    \end{tabular}
    \caption{Comparison with character-based neural models on the positionally capitalized Wikipedia text simplification data set.}
    \label{tab:wiki_simpl}
\end{table*}

\subsection{Large-scale, Position-Invariant Comparison}
\label{sec:large_scale}
\begin{table*}[htb]
    \centering
    \begin{tabular}{rl|c|c|c|r|r}
    \multicolumn{2}{c}{\textbf{System}} & \textbf{Precision} & \textbf{Recall} & \textbf{F1} & \textbf{Speed} & \textbf{\# of params}\\
    \hline
    \multirow{2}{*}{\textbf{5-gram FST}} & 
    large & 86.39 & 38.60 & 53.36 & 1x & 30M\\
    &pruned & 85.87 & 37.71 & 52.41 & 88x & 1M \\
    \hline
    \multirow{4}{*}{\textbf{char RNN}}
    &small, 1-layer uni-, dec. & 69.11 & 22.86 & 34.35 & 0.7x & 230K \\
    &small, 1-layer bi-, enc. \& dec. & 86.12 & 75.07 & 80.22 & 0.5x & 400K\\
    &small, 2-layer bi-, end. \& dec. & 85.57 & 72.05 & 77.48 & 0.4x & 620K\\
    &large, 2-layer bi-, enc. \& dec. & \textbf{87.06} & 78.09 & 82.33 & 0.1x & 8.4M\\
    &large, 3-layer bi-, enc. \& dec. & 86.43 & 77.76 & 81.87 & 0.05x & 11.7M\\
    \hline
    \multirow{1}{*}{\textbf{this work}} & small, 1-layer bi-dir., two-level & 86.95  & \textbf{79.81} & \textbf{83.23} & 2.2x & 1.3M
    \end{tabular}
    \caption{Large-scale comparison across model types on intrinsically capitalized Wikipedia edit history data set.}
    \label{tab:wiki_edit}
\end{table*}

The comparison in Section~\ref{sec:char_wiki_simpl} focuses on the accuracy of character-based models with different architectures. However, there are some limitations of this study. First, the training data is small and there is no comparison in terms of relative speed and model size. Also, the evaluation metrics are inflated because of positional capitalization. In this section, we scale up various models to billions of training examples and report results in terms of non-positional non-lowercase F1. In this new setting, we use a large data set of digitized books and newswire articles. The data set has 1.5 billion sentences. We use 90\% of the data for training and sample a subset from the remaining 10\% for validation. The testing is on the out-of-domain Wikipedia edit history data with intrinsic capitalization as described in Section~\ref{sec:wiki_edit_data}.

We present three groups of results in Table~\ref{tab:wiki_edit}. The FST group represents word-based HMM models with a wide coverage of mixed case words. The character-based RNN group is our implementation of  \newcite{susanto-etal-2016-learning} with various hyper-parameters. The third group is our hierarchical word- and character-based RNN.

For training our model, we first train a teacher model (second column in Table~\ref{tab:network_specs}) on the training set until the sentence error rate converges on the validation set. We next generate a new training set by running inference with the teacher model and a sentence prefix \texttt{so,}. The sentence prefix is removed in the decoded output. The regenerated data is used to train all of the neural models in a hard EM fashion. All neural models in Table~\ref{tab:wiki_edit} are taught by the same teacher model. The FST models use the inference-time prefix trick \cite{6638968} to infer the truecase. We note that the FST model is trained on a different large data set with roughly 10 million sentences and is therefore not directly comparable to our model. Our goal of comparing with FSTs is to demonstrate how our neural model compares with a large-scale FST model in terms of speed and model size. \newcite{susanto-etal-2016-learning} have shown that HMM models are worse than character-based RNNs. We here focus on comparing accuracy with character-based RNNs.

We report CPU speed relative to the large FST model. We use batch size of 1 at inference time for all models. All models are implemented in TensorFlow and quantized after training. We compare the model sizes in terms of the total number of parameters. 

FST models have high precision but low recall, indicating the coverage problem of word-based models. The one-layer uni-directional character-based RNN scores low in both precision and recall. Multi-layer bi-directional character based models show strong results only when using at least three layers, including the decoder layer. However, it is clear that even a single-layer bidirectional character-based RNN is already 4 times slower than a single-layer word- and character-based hierarchical RNN. Using two layers in both the encoder and the decoder, with about the same F1 score as the hierarchical one, a purely character-based model is nearly 20 times slower than a hierarchical one.


\subsection{First Word Accuracy}
\label{sec:distill_prefixes}
In this section, we study the effectiveness of the distillation method for recovering truecasing of sentence-initial words. We vary the sentence prefix we use in sequence distillation as well as the optional sentence prefix to use at inference time. Instead of F1, we focus on the accuracy of the sentence-initial words.
Table~\ref{tab:first_words} shows the impact of the prefix choices. First, if we only perform standard sequence distillation, the student model learns from the teacher model to do positional truecasing. As a result, it has a strong tendency to turn sentence-initial words to first letter uppercase, leading to low accuracy (16.08\%). We can see that the inference time prefix is very effective. Using either the short prefix \texttt{so,} or the long prefix \texttt{it is known that}, the accuracy jumps to over 89\%. But it comes with an inference time cost of decoding a longer input sentence. The distillation prefix, however, renders the inference prefix unnecessary to a large extent (89.25\% versus 89.67\% and 90.17\% versus 89.92\%). The inference time cost has been transferred to the training time. There does not seem to be a large difference in performance between long versus short distillation prefixes, with the caveat that only a small set of prefixes were explored.

\begin{table}[htb]
    \centering
    \begin{tabular}{c|c|c}
    \textbf{Distill. Prefix} & \textbf{Infer. Prefix} & \textbf{Acc.} \\
    \hline
          \multirow{3}{*}{\small{n/a}} &  \small{n/a} & 16.08 \\
                               &  \texttt{\small{so,}} & 89.58 \\
                               &  \texttt{\small{it is known that}} & 89.25 \\
    \hline
         \multirow{3}{*}{\texttt{\small{so,}}} &  \small{n/a} & 89.25 \\
                                      &  \texttt{\small{so,}} & 89.58 \\ 
                                      &  \texttt{\small{it is known that}} & 89.67 \\ 
    \hline
         \multirow{3}{*}{\texttt{\small{it is known that}}} & \small{n/a} & \textbf{90.17} \\
                                                    &  \texttt{\small{so,}} & 89.67 \\ 
                                                    &  \texttt{\small{it is known that}} & 89.92 \\ 
    \end{tabular}
    \caption{First word accuracy with various distillation and inference sentence prefixes.}
    \label{tab:first_words}
\end{table}

\subsection{Impact of Training Data Size}
\begin{figure}
    \centering
    \includegraphics[width=\columnwidth]{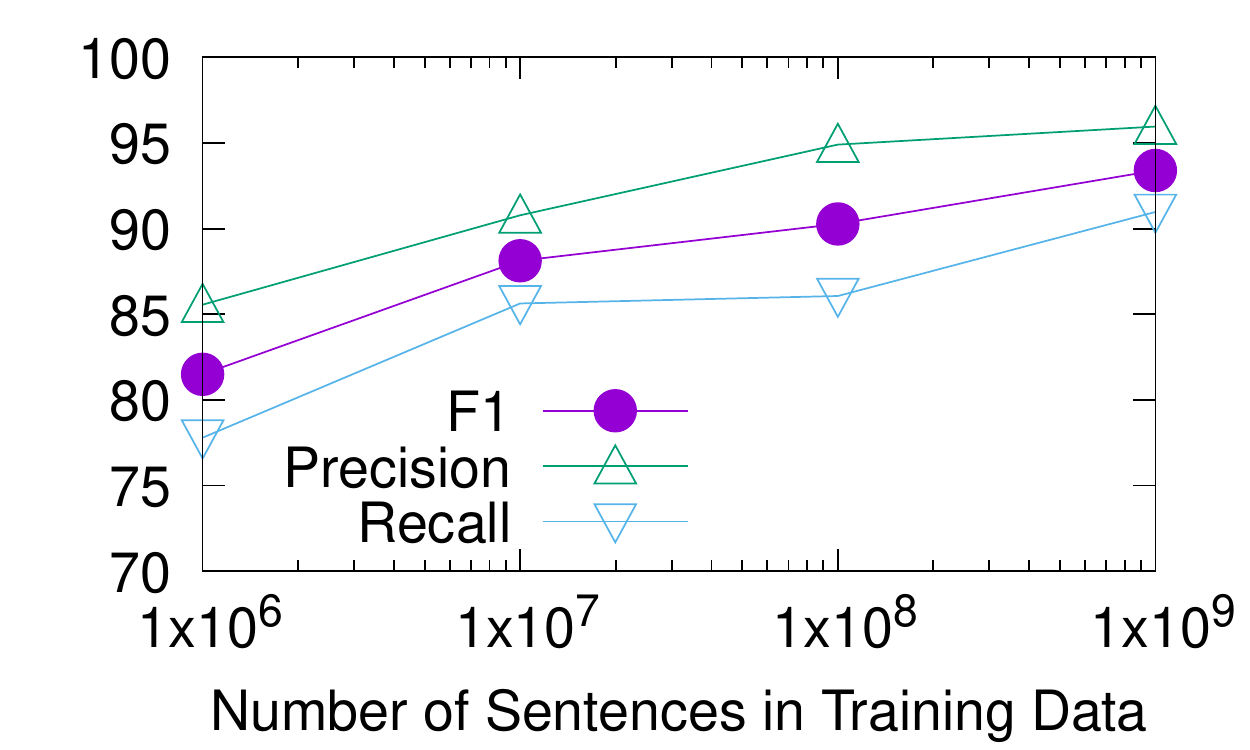} 
    \caption{F1 scores as a function of training data size.}
    \label{fig:f1_plot}
\end{figure}

In this section, we study the correlation between training data size and model accuracy. In Figure~\ref{fig:f1_plot}, we plot Precision/Recall/F1 versus the number of training sentences. The model in the plot is the teacher model trained on the original (books and news) data with various down-sampling rates (1\%, 0.1\%, 0.01\%). The results are non-lowercase F1 on the validation set of the Wikipedia simplification data set in Section~\ref{sec:char_wiki_simpl}. It is noteworthy that with $10^{9}$ training sentences, the F1 score (93.4\%) on the validation set matches the best model trained on the in-domain Wikipedia simplification data. It is to be seen how the model performs on a domain that is drastically different from online publications, for example, social media and noisy text such as ASR transcripts.


\subsection{Error Analysis}
Finally, we look at the wins and losses on the Wikipedia edit history test set for qualitative analysis.
\begin{table*}[htb!]
    \centering
    \begin{tabular}{l|l}
          &\textbf{Wins} \\
    \hline
    & \small{\texttt{in 2009 \textbf{LiveMint} (online news portal rated GECT among the top 50 ...}}\\
    & \small{\texttt{\textbf{Palm} launched \textbf{WebOS} in January 2009, then called \textbf{Palm} \textbf{WebOS} ...}}\\
    & \small{\texttt{\textbf{... WinZip-compatible} AES encryption ...}}\\
    & \small{\texttt{\textbf{LEDs} have also been developed that use several quantum wells to emit light ...}}\\
    \end{tabular}
    \caption{Wins of mixed case words.}
    \label{tab:wins}
\end{table*}
\begin{table*}[htb!]
    \centering
    \begin{tabular}{l|l}
    &\textbf{Losses} \\
    \hline
     \multirow{5}{*}{\textbf{Reference is wrong}}
     & \small{\texttt{Des Connely a catholic priest ...}} \\
     & \small{\texttt{Des Connely a \textbf{Catholic} priest ...}} \\
     \\
     & \small{\texttt{miller studied electrical engineering and computer science at ...}} \\
     & \small{\texttt{\textbf{Miller} studied electrical engineering and computer science at ...}} \\
     \\
      \multirow{5}{*}{\textbf{Under-capitalization}}
      & \small{\texttt{Holungen was largely unaffected by direct action of the two World Wars.}} \\
      & \small{\texttt{Holungen was largely unaffected by direct action of the two \textbf{world wars}.}} \\
      \\
      & \small{\texttt{the Scripps Aviary is home to many colorful birds}} \\
      & \small{\texttt{the Scripps \textbf{aviary} is home to many colorful birds}} \\ 
      \\
      \multirow{3}{*}{\textbf{Over-capitalization}}
      & \small{\texttt{the college chapel was built after the First World War.}} \\
      & \small{\texttt{the \textbf{College Chapel} was built after the First World War.}} \\
      \\
      \multirow{5}{*}{\textbf{Ambiguous}}
      & \small{\texttt{the Open-Air Sculpture Gallery contains 93 works by 77 artists.}} \\
      & \small{\texttt{the \textbf{open-air sculpture gallery} contains 93 works by 77 artists.}} \\
      \\
      & \small{\texttt{... which is heart of IT in Bangalore.}} \\
      & \small{\texttt{... which is heart of \textbf{it} in Bangalore.}} \\
      \\
      \multirow{4}{*}{\textbf{Uncommon mixed case}}
      & \small{\texttt{Magellan was purchased by MiTAC on December 15th 2008.}} \\
     & \small{\texttt{Magellan was purchased by \textbf{MITAC} on December 15th 2008.}} \\
     \\
     & \small{\texttt{text-based codes use OCR software for mTicket.}} \\
    & \small{\texttt{text-based codes use OCR software for \textbf{Mticket}.}} \\
    \end{tabular}
    \caption{Categories of losses. In each pair of sentences, the top one is the reference and the one below is the hypothesis.}
    \label{tab:error_analysis}
\end{table*}
Table~\ref{tab:wins} lists the wins that highlight the strength of the model in dealing with a variety of mixed case words within context. For example, the word \texttt{palm} has multiple senses. In the entity name \texttt{Palm WebOS}, it is taking a new word sense of being a commercial brand name. The model consistently predicted correct capitalization for both instances of \texttt{Palm}. The word \texttt{LEDs} shows that the class of mixed words is rich with inflection. It is important to look at the following word \texttt{have} to infer the correct form for the input word \texttt{leds}. \texttt{WinZip-compatible} is another example of mixed case words that are formed through word compounding.

Table~\ref{tab:error_analysis} categorizes the losses of our model. First, the references are not perfect. We have cases where the model is actually correct. For example, it should be \texttt{a Catholic priest}, not \texttt{a catholic priest}. Among errors that are either due to under-capitalization or over-capitalization, some can arguably have alternative interpretations. But for humans with more world knowledge, it is more clear which casing is more likely and common. For example, \texttt{the two world wars} is acceptable syntactically. But given that in most contexts, the reading should be \texttt{the two World Wars}, referring to the modern history. At the same time, the model is able to get \texttt{the First World War} correct in most cases. Improving the consistency of the model across contexts is a very interesting future research direction. It is also important for downstream NLP applications. Finally, the class of mixed case words has a long tail distribution. Predicting \texttt{MiTAC} and \texttt{mTicket} correctly probably requires training on a knowledge data base with lots of mixed case words.

\section{Conclusion}
We proposed a dual-granularity RNN model for Truecasing with the efficiency of word-level models and the generalizability of character-level models. We systematically compared our approach to the state-of-the-art models on a small public data set. We also demonstrated superior results in terms of efficiency-vs-accuracy trade-offs when our model was scaled up to billions of training examples. Our partial sequence distillation method provided a simple and effective solution to the problem of recovering true casing for sentence-initial words. 

\clearpage

\bibliography{emnlp2021}
\bibliographystyle{acl_natbib}



\end{document}